\documentclass{article}
\usepackage{spconf,amsmath,graphicx,amsfonts, bbm}
\usepackage{algorithmic}
\usepackage{algorithm}
\usepackage{url}

\title{Optimal Transport-based Domain Alignment as a Preprocessing Step for Federated Learning\footnote{
}}
\name{Luiz Manella Pereira and M. Hadi Amini, \textit{Senior Member, IEEE }\thanks{This work has been accepted to IEEE ICIP 2025. \textcopyright 2025 IEEE. Personal use of this material is permitted. Permission from IEEE must be obtained for all other uses. 
\\
This work is based upon the work supported by the National Center for Transportation Cybersecurity and Resiliency (TraCR) (a U.S. Department of Transportation National University Transportation Center) headquartered at Clemson University, Clemson, South Carolina, USA. Any opinions, findings, conclusions, and recommendations expressed in this material are those of the author(s) and do not necessarily reflect the views of TraCR, and the U.S. Government assumes no liability for the contents or use thereof. This material is also partly based upon Luiz Manella Pereira's work supported by the Graduate Assistantships in Areas of National Need (GAANN) fellowship. 
}}
\address{Knight Foundation School of Computing and Information Sciences, Florida International University\\  Security, Optimization, and Learning for InterDependent networks laboratory (solid lab)\\www.solidlab.network}

\begin{document}
%
\maketitle
\begin{abstract}
Federated learning (FL) is a subfield of machine learning that avoids sharing local data with a central server, which can enhance privacy and scalability. The inability to consolidate data leads to a unique problem called dataset imbalance, where agents in a network do not have equal representation of the labels one is trying to learn to predict. In FL, fusing locally-trained models with unbalanced datasets may deteriorate the performance of global model aggregation, and reduce the quality of updated local models and the accuracy of the distributed agents' decisions. In this work, we introduce an Optimal Transport-based preprocessing algorithm that aligns the datasets by minimizing the distributional discrepancy of data along the edge devices. We accomplish this by leveraging Wasserstein barycenters when computing channel-wise averages. These barycenters are collected in a trusted central server where they collectively generate a target RGB space. By projecting our dataset towards this target space, we minimize the distributional discrepancy on a global level, which facilitates the learning process due to a minimization of variance across the samples. We demonstrate the capabilities of the proposed approach over the CIFAR-10 dataset, where we show its capability of reaching higher degrees of generalization in fewer communication rounds.
\end{abstract}
\section{Introduction}
\label{sec:intro}

Federated Learning (FL), a subdomain of Machine Learning (ML), addresses the challenges of decentralized or distributed model training. It offers a compelling framework for scenarios in which data cannot be centrally aggregated due to privacy constraints, thereby promoting compliance with data protection regulations and enhancing scalability \cite{mcmahan2017communication}. Beyond its foundational role in privacy-preserving learning, FL also facilitates model personalization—adapting learning outcomes to individual users across the network—an increasingly relevant objective given the heterogeneity of user behavior and datasets. A comprehensive overview of the challenges and practical implementations of personalized federated learning is presented in \cite{tan2022towards}.

Despite its broad applicability, particularly in contexts with stringent data privacy constraints, FL introduces a set of constraints that must be carefully addressed to ensure robust and efficient model training. These constraints include limited communication bandwidth, restricted computation at edge devices, privacy preservation requirements, and data heterogeneity and imbalance. Dataset imbalance in FL emerges when edge devices possess non-uniform class distributions, disparate dataset sizes, or varying data quality \cite{li2020federatedchallenges, wu2023fedprof}. In this work, we propose a preprocessing framework that addresses this imbalance challenge in a model- and algorithm-agnostic manner. Our method aligns and transforms local datasets into a shared representation space that captures statistical information from all participating agents in the network.

\section{Related Works}
The issue of dataset imbalance, also known as the domain-distribution discrepancy problem or the multiple-source domain problem, has been a core challenge in federated learning. In this section, we will review various approaches to address this challenge. Our approach is, to our knowledge, the first to perform zero-shot distribution alignment in FL and only the second to incorporate optimal transport (OT). Unlike prior work, our method eliminates the need for iterative training-time computation by applying OT as a preprocessing step, which reduces overhead and enhances scalability. Although it may be interesting to think of various preprocessing techniques to compare against, to our knowledge, none exist. In our research, we have only seen local alignments and processing techniques such as feature normalization (e.g., min-max scaling) or standardization (e.g., Z-score scaling). While these techniques are useful to help train models by scaling feature values to desirable ranges, they do not align samples across agents. The lack of literature further supports the importance of our idea.

The most relatable work we have found is that of Farnia et al. \cite{farnia2022optimal}, where the authors introduce an OT-based alignment step during the process of training the model. While their method learns the space they project to and the projection map dynamically during training, our approach performs alignment before training, avoiding iterative optimization. The authors begin by extending the standard OT task between two distributions to a multi-marginal OT problem. They then use these results to create a min-max optimization problem, which leads to their algorithm called \textit{FedOT}, which is a dynamic approach to learning the data-transformation map. In this work, we instead present a zero-shot alignment that leverages \cite{ferradans2014regularized} to create the alignment map, and in turn, lower computation cost.

In \cite{wang2022generalizing}, Wang et al. comprehensively surveyed the field of domain adaptation. They describe three primary techniques used to solve this problem: data manipulation, representation learning, and learning strategy. While domain adaptation and generalization have been broadly studied, only a few methods directly address distribution alignment in FL. These include transfer learning approaches \cite{chen2020fedhealth}, representation aggregation \cite{wu2021collaborative}, and adversarial training \cite{zhang2021federated}. However, these either require task-specific models or exhibit inferior performance compared to OT-based methods like L2A-OT \cite{zhou2020learning}. First, \cite{chen2020fedhealth} focuses on personalization through transfer learning. The pseudo domain generalization occurs in the transfer step, as the \textit{FedHealth} algorithm aligns the lower levels of the convolutional neural network (CNN) aggregated in the cloud with those locally stored, and personalized, in each agent's device. On the other hand, in \cite{wu2021collaborative}, Wu and Gong tackle the learning strategy approach of domain generalization (DG) and unsupervised domain adaptation (UDA). In their algorithm, called Collaborative OPtimization and Aggregation (COPA), each agent in the network contains a local dataset, a feature extractor, and a classifier. The local feature extractors are collected to create a global feature extractor that is domain-invariant. The local classifiers are ensembled to create the global classifier. These global models are used to update the local models until convergence. Lastly, \cite{zhang2021federated} subscribes to the ``learning strategy'' approach to solving domain generalization. They accomplish this through their algorithm \textit{FedADG}, which uses an adversarial component to measure and then align different source distributions to a globally known, shared distribution. While \textit{FedADG} improves on some test cases, the authors mention their approach to generating invariant features yields results that are typically worse than an OT-based approach called L2A-OT \cite{zhou2020learning}. 

Thus far, we have covered papers that are directly related to our work. We now focus on other research that indirectly tackles, or at least provides, a direction to work with imbalanced datasets. We first turn to clustering in federated learning. In \cite{sattler2020clustered}, Sattler et al. propose their main algorithm, clustered federated learning (CFL). In an imbalanced dataset paradigm, the agents' data-generating distributions are different. To overcome this challenge, CFL clusters the incoming gradients of the agents through a recursive cosine similarity-based bipartitioning and then updates the model's parameters once for each cluster by averaging the gradients of the clients in each cluster respectively. Moreover, in \cite{hsu2020federated}, Hsu et al. create a custom dataset and analyze the degree of imbalance based on geo-location tagging and distributional statistics. To tackle the class distribution shift problem in FL, they began with, \textit{FedAvg}, short for federated averaging \cite{mcmahan2017communication}, and updated it to include variance mitigation and a resampling strategy via adding weights to each agent. Their algorithms, respectively called \textit{FedVC} and \textit{FedIR}, computed the weights as a ratio of the target joint distribution of images and class labels and a sampled joint distribution of a specific client's image-label pair. The authors propose using Bayes theorem to reformulate the weight function as a ratio of the global distribution of class $y$ and the agent's local distribution of class $y$. The weights are used to scale the updates of the global model sitting in the trusted server and lead to better accuracy. 

Lastly, personalization is an interesting direction because it has been both used as a solution to the aforementioned problem and as an application to improve local, customized solutions for FL. Fallah et al. present \textit{MAML} an algorithm that trains an ``initial shared model'' which is updated using a few gradient steps to yield a good, personalized model \cite{fallah2020personalized}. For more algorithms regarding personalization, we refer readers to \cite{tan2022towards}. n summary, while prior work has explored various strategies—dynamic OT, adversarial alignment, clustering, and personalization—none provide a model-agnostic, zero-shot distribution alignment mechanism. Our method fills this gap with a computationally efficient, one-time preprocessing step that enhances the stability of training.

\section{Preliminaries}
\subsection{Federated Learning}
The goal of federated learning (FL) is to train a single (global) model to make accurate predictions across all agents in a network. FL problems contain constraints, including computational power at the edge, communication bandwidth, and data heterogeneity. Privacy is one of the key motivations for using FL which prevents the amalgamation of data in a central server. To overcome this problem, FL research introduced novel techniques that are variations of parameter averaging or gradient averaging. A more comprehensive look at the challenges of FL and associated solutions can be found in \cite{li2020federatedchallenges}. Independently of the approach, at its core, FL looks for a solution to 
\begin{equation}\label{fl-equation}
    \min_w f(w), \quad f(w) = \sum_{a=1}^A p_a f_a(w)
\end{equation}
where $A$ is the number of devices in our network, $p_a \geq 0$, and $ \sum_a p_a = 1$. Each agent has a local objective, $f_a(w)$, which is averaged together as the global objective $f(w)$. For more details into equation \ref{fl-equation}, we refer readers to \cite{li2020federatedchallenges}.

\subsection{Optimal Transport}
Optimal Transport (OT), founded by Gaspard Monge, is a mathematical framework that addresses the problem of finding the most cost-efficient way to transport mass from a source distribution to a target distribution \cite{monge1781memoire}. OT has found widespread applications in various fields, including economics, computer vision, machine learning, and statistics. At its core, OT seeks to quantify the discrepancy between two probability distributions by defining a distance metric that considers both the magnitudes of the masses being transported and the distances over which they are moved. Unlike traditional distance metrics, such as Euclidean distance, which focus solely on point-to-point comparisons, OT provides a more geometrically nuanced measure that captures the structural similarities between distributions. 

Kantorovich's formulation of OT introduces the concept of mass-splitting, which allows mass from the domain to be broken up and mapped to different locations in the codomain. Mathematically, it reads:
\begin{equation}
    L_\textbf{C}(\textbf{a},\textbf{b}) = \min_{P\in U(a,b)}\langle \textbf{C},\textbf{P} \rangle = \sum_{i,j} \textbf{C}_{i,j}\textbf{P}_{i,j},
\end{equation}
given the set of \textit{admissible couplings}
\begin{equation*}
   U(\textbf{a},\textbf{b}) = \bigg\{ P \in R_+^{n\times m} : \textbf{P}\mathbbm{1}_m = \textbf{a}, \textbf{P}^T \mathbbm{1}_n = \textbf{b} \bigg\}
\end{equation*}
Intuitively, one is looking for a permutation matrix $\textbf{P}$ that determines how to distribute mass in a cost-minimizing fashion given the transportation cost $\textbf{C}$.

OT presents a novel approach to comparing two probability distributions. With additional constraints, OT yields a metric, or a distance function, called the \textit{Wasserstein metric}, or \textit{Earth Mover's Distance}. Suppose that for some $p\geq 1$ and $\textbf{C}=\textbf{D}^p$ where $\textbf{D}\in \mathbb{R}^{n \times n}$ is a distance on $[\![ n ]\!]$, the following hold:
1- $\textbf{D}$ is symmetric,
    2- \text{diag}(\textbf{D}) = $\mathbf{0}$, and 3-  $\forall (i,j,k) \in [\![ n ]\!]^3, \textbf{D}_{i,k} \leq \textbf{D}_{i,j} + \textbf{D}_{j,k}$. 
 Then the p-Wasserstein distance is
\begin{equation}
    W_p(\textbf{a},\textbf{b})= L_{\textbf{D}^p}(\textbf{a},\textbf{b})^{1/p}.
\end{equation}

Equipped with a distance function, we can now define an averaging function, called a Wasserstein barycenter (WB): 

\begin{equation}\label{wasserstein_barycenter}
    \min_{\textbf{a}\in\Sigma_n}\sum_{s=1}^S\lambda_s W_p^p(\textbf{a}, \textbf{b}_s),
\end{equation}

\noindent where $\lambda_s$ is a real-valued weight (usually a uniform distribution such that each input probability vector is given an equal amount of importance \cite{cuturi2014fast}) and $\Sigma_n$ is a probability simplex with $n$ bins.
Intuitively, given a set of input probability vectors $\mathbf{b}_s$, we are looking for a probability vector $\mathbf{a}$ that minimizes the weighted sum of the p-Wasserstein distance between $\mathbf{a}$ and each $\mathbf{b}$. 

There are different methods to solve the optimal transport problem and to compute Wasserstein barycenters. The current state-of-the-art methods rely on entropic regularization. In \cite{cuturi2013sinkhorn}, Marco Cuturi introduces a solution to quickly solve the entropy regularized OT problem, $W_{reg}(\textbf{a}, \textbf{b})$, using Sinkhorn's algorithm, which now reads: For $\lambda > 0$,
\begin{equation}
    d^\lambda_M (a,b) = \langle P^\lambda, C \rangle, 
\end{equation}
where
$ P^\lambda = \text{argmin}_{P\in U(a,b)}\langle P,C \rangle - \frac{1}{\lambda}h(P).$    
Equipped with a computational solver, Benamou et al. leveraged iterative Bregman projections to compute the entropic regularized Wasserstein barycenter, which can be seen in \cite{benamou2015iterative}, leading to a faster and more general solution. The entropic regularized barycenter problem can be written as an extension of equation \ref{wasserstein_barycenter}:
\begin{equation}\label{regularized_wasserstein_barycenter}
     \min_{\mathbf{a}\in\Sigma_n}\sum_{s=1}^S\lambda_s W_{reg}(\mathbf{a}, \mathbf{b}_s)
\end{equation}
where $\boldsymbol{\lambda} = \{\lambda_s \}_{s=1}^S \in \Sigma_S$.

\section{Optimal Transport-based Preprocessing}
In this section, we introduce the preprocessing step that minimizes the distributional discrepancy in our network. We achieve this distribution-alignment goal by generating a target space to which we project all local data. Our proposed method has two main steps: the creation of the target space, and the projection step to perform alignment.

To create a relevant target space to align images to, it needs to contain information from all the agents in the network. We accomplish this in a two-step fashion. First, we compute representations of the local data through a channel-wise Wasserstein barycenter of the local images. The approach requires separating each local image its three color channels, grouping them by the respective color channels, and lastly computing the WB for each channel. In this work, we are using colored images, therefore, channel-wise implies red, green, and blue channels. The local computations produce an RGB-triplet called the local WBs. The second step is to generate the target space by aggregating all local WBs in a central server, and repeating the same process of computing channel-wise barycenters; this yields the RGB-tripled called the global WB, or our target space. Lastly, we broadcast the global barycenters to the agents in the network and then align the local images to the target space by projecting them to the global WB. The projection process is composed of computing transportation plans to the target space that allows the color channels of the original images to be transferred and aligned, similarly to computing color transfer maps or domain adaptation in \cite{ferradans2014regularized} respectively. The entire preprocessing steps are explicitly laid out in algorithm \ref{alg:ot-preprocess}. For clarity, $\mathcal{WB}^a_{r,g,b}$ are the Wasserstein barycenters of agent $a$ and the respective color channel, $\mathbf{Img}^{red, green, blue}_i$ is a specific color channel of the $i^{th}$ image, and $\mathcal{WB}^G$ is the target space (the global Wasserstein barycenter) composed of the global RGB-triplet.

\begin{algorithm}
    \caption{OT-based preprocessing}
    \vskip 0.1cm
    \textbf{Definitions:} Let $\mathbf{B}=\{ \mathbf{b}_s \}_{s=1}^S$ such that $\text{WB}(\mathbf{B})$ is the solution to equation (\ref{regularized_wasserstein_barycenter}). 
    \vskip 0.1cm
    \hrule
    \vskip 0.1cm
    \label{alg:ot-preprocess}
    \begin{algorithmic}
        \STATE \textbf{For each agent $a=1,2,\ldots,N$}
            \STATE \hspace{0.5cm} $\mathbf{R} = \{ \mathbf{Img}^{red}_i \}_{i=1}^M$
            \STATE \hspace{0.5cm} $\mathbf{G} = \{ \mathbf{Img}^{green}_i \}_{i=1}^M$
            \STATE \hspace{0.5cm} $\mathbf{B} = \{ \mathbf{Img}^{blue}_i \}_{i=1}^M$
            \STATE \hspace{0.5cm} $\mathcal{WB}^a_r = \text{WB}(\mathbf{R})$
            \STATE \hspace{0.5cm} $\mathcal{WB}^a_g =\text{WB}(\mathbf{G})$
            \STATE \hspace{0.5cm} $\mathcal{WB}^a_b = \text{WB}(\mathbf{B})$
            \STATE \hspace{0.5cm} Distribute $\mathcal{WB}^a_r$, $\mathcal{W}^a_g$, $\mathcal{WB}^a_b$ to a central server
        \STATE $\mathbf{R}_G = \{ \mathcal{WB}_r^a \}_{a=1}^N$
        \STATE $\mathbf{G}_G = \{ \mathcal{WB}_g^a \}_{a=1}^N$
        \STATE $\mathbf{B}_G = \{ \mathcal{WB}_b^a \}_{a=1}^N$
        \STATE $\mathcal{WB}^G_r = \text{WB}(\mathbf{R}_G)$
        \STATE $\mathcal{WB}^G_g = \text{WB}(\mathbf{G}_G)$
        \STATE $\mathcal{WB}^G_b = \text{WB}(\mathbf{B}_G)$
        \STATE $\mathcal{WB}^G = \{\mathcal{W}^G_r, \mathcal{WB}^G_g, \mathcal{WB}^G_b\}$
        \STATE Distribute $\mathcal{WB}^G$ to \textbf{all} agents
        \STATE \textbf{For each agent $a=1,2,\ldots,N$}
            \STATE \hspace{0.5cm} \textbf{For each image $i=1,2,\ldots,M$ of agent $a$}
                \STATE \hspace{1cm} Project image $i \rightarrow \mathcal{WB}^G$
    \end{algorithmic}
\end{algorithm}

\section{Experiments and Results}
Our framework is built independently of the learning algorithm, which allows for flexible integration into other pipelines that may have different goals. To demonstrate the advantages of using our preprocessing step, we require a learning algorithm. To this end, we chose to work with federated averaging, \textit{FedAvg}, which trains local algorithms and performs parameter averaging to aggregate the local models into a global model. To show the improvements our method provides to the learning process, we designed a few experiments over the CIFAR-10 dataset. Each experiment begins with the following steps. First, we design the network based on a desired number of agents. Next, we initialize their models identically to avoid degeneracy \cite{mcmahan2017communication}. We use both a custom CNN and ResNet9 in our experiments. Next, we distribute the data by uniformly sampling them without replacement. Lastly, we align the data using our preprocessing algorithm (\ref{alg:ot-preprocess}) and begin training. 

There are various ways of deciding when to perform the synchronization step, also called communication round, which is where parameter averaging happens. The trade-off between choosing the number of local epochs and communication rounds when training a global model is dependent on how many agents participate in the model aggregation step. As per convention, $N$ is the total number of agents and $P$ is the number of participating agents. In experiments where $N=P$, we used two local training epochs per synchronization step. When $P<<N$, the number of samples in the local dataset becomes closer to the batch size and we therefore increase the number of epochs to five. The results of our simulations using the custom CNN and the ResNet9 model are shown, respectively, in Tables \ref{tab:accuracy-results} and \ref{tab:resnet-accuracy-results}. Moreover, Table \ref{tab:accuracy-comparison} shows a comparison of our algorithm with other approaches. 

Our OT-preprocessing algorithm, paired with \textit{FedAvg}, an algorithm known for its simplicity, yielded the best results. Our simulations, while not using the exact same hyperparameters, are undoubtedly comparable since we achieved higher accuracy scores with a simpler, and smaller, model. These facts open the door for merging our work with algorithms that are tailored for other applications. 

\begin{table}
\caption{Comparison between \textit{FedAvg} (i) with our preprocessing (column 3); (ii) without it (column 4). The first column is  number of clients in the simulated network (N) and the number of clients used to train the global model (P). The second column (Communication Rounds), is the number of synchronization steps.\\\label{tab:accuracy-results}}
\centering
\begin{tabular}{|c|c|c|c|c|}
\hline
Client (N/P) &  Comm. Rounds & \textbf{Our Work} & \textit{FedAvg}\\
\hline
5/5 & 35 & \textbf{99.62} & 71.22\\
\hline
10/10 & 70 &  \textbf{99.33} & 70.48\\
\hline
20/20 & 200 & \textbf{99.36} & 69.89\\
\hline
50/50 & 250 & \textbf{97.84} & 69.73\\
\hline
100/100 & 500 & \textbf{94.95} & 65.01\\
\hline
10/5 & 100 & \textbf{99.85} & 71.34\\
\hline
20/10 & 150 & \textbf{97.5} & 65.66\\
\hline
50/10 & 500 & \textbf{95.04} & 66.45\\
\hline
100/10 & 1000 & \textbf{93.34} & 66.16\\
\hline
\end{tabular}
\end{table}

\begin{table}
\caption{This table contains the results of simulations using our approach with a ResNet9 model and our custom CNN (for comparison). For clarification, ``Comm. Rounds'' is short for ``Communication Rounds.''\\\label{tab:resnet-accuracy-results}}
\centering
\begin{tabular}{|c|c|c|c|c|}
\hline
& \multicolumn{2}{c|}{Comm. Rounds} & \multicolumn{2}{c|}{Testing Accuracy}\\
\hline
Client (N/P) &  ResNet9 & CNN &  ResNet9 & CNN\\
\hline
5/5 & 554 & 35 & 98.61 & 99.62\\
\hline
10/10 & 200 & 70 &  95.45 & 99.33\\
\hline
20/20 & 200 & 200 &  92.13 & 99.36\\
\hline
\end{tabular}
\end{table}

\begin{table}[t]
\caption{This table compares the results of various approaches to federated learning. The first two rows are direct results from our experiments. The subsequent rows are results obtained from the respective papers. Readers should look at the cited work for implementation details which lead to the varying accuracy scores for the same algorithms (as can be seen below). For further clarification, rows 1 and 2 (referencing our Algorithm and \textit{FedAvg}), are simulations with N=100 and P=10.\\\label{tab:accuracy-comparison}}
\centering
\begin{tabular}{|c|c|c|}
\hline
\textbf{Reference} & \textbf{Method} & \textbf{Accuracy}\\
\hline
Our work & Algorithm \ref{alg:ot-preprocess} & \textbf{93.34} \\
\hline
\cite{mcmahan2017communication} & \textit{FedAvg} & 66.16 \\
\hline\hline
\cite{li2021model} & MOON & 69.1 \\
\hline
\cite{li2021model} & FedAvg & 66.3 \\
\hline
\cite{li2021model} & FedProx & 66.9 \\
\hline
\cite{li2021model} & SCAFFOLD & 66.6 \\
\hline
\cite{li2021model} & SOLO & 46.3 \\
\hline\hline
\cite{wang2020federated} & FedMA & 87.53 \\
\hline
\cite{wang2020federated} & FedProx & 85.32 \\
\hline
\cite{wang2020federated} & FedAvg & 86.29 \\
\hline\hline
\cite{luo2021no} & FedAvg (CCVR) & 71.03\\
\hline
\cite{luo2021no} & FedProx (CCVR) & 70.99\\
\hline
\cite{luo2021no} & FedAvgM (CCVR) & 71.49\\
\hline
\cite{luo2021no} & MOON (CCVR) & 71.29\\
\hline\hline
\cite{farnia2022optimal} & FedOT & 72.2\\
\hline
\end{tabular}
\end{table}

\section{Implementation Details}\label{implementation-details}
In this short section, we will clarify details for reproducibility. The custom CNN has an input convolutional layer (conv layer) with a $5\times 5$ kernel and 64 filters and its output is fed into a max pool layer with a $2\times 2$ kernel. Next, we have one more conv layer with the same kernel size but with 128 filters. The output is fed into a series of 3 linear layer, whose respective output dimensions are 256, 128, and 10. When training either CNN or ResNet9, we use a batch size of 16 with an Adam optimizer and a learning rate of $1e{-3}$.

The core of our algorithm relies on efficiently computing an entropic-regularized transport and a Wasserstein barycenter efficiently . We leverage the Python Optimal Transport (POT) package \footnote{\url{https://pythonot.github.io/}} which contains various solvers; for the transportation problem, we use the Sinkhorn algorithm, while for the barycenter we use the Bregmen projection method. Both methods require a regularization term, for which we use $1e-2$ and $1e-1$ respectively. Furthermore, a trick used to reduce the computations required is to select subsets of pixels from the domain and codomain. In our experiments, we uniformly sample a subset of 250 pixels and use them as input when computing barycenters and performing projections. 

\section{Complexity Analysis}\label{complexity_analysis_section}
An important tradeoff to consider with our algorithm is the additional time that must be paid to convert the original dataset into an aligned dataset. As our method does not affect the learning algorithm, we only analyze the complexity of computing the barycenters and projecting the local data to the global barycenter, which is done prior to training. Therefore, this is an additional cost paid in addition to training. First, \cite{kroshnin2019complexity} gives us a complexity analysis for the iterative Bregman projection method of computing regularized barycenters, as introduced by \cite{benamou2015iterative}. Given $n$ samples of dimension $d$ and regularization parameter $\epsilon$, we have a complexity of $O(nd^2/\epsilon^2)$. Moreover, the projection of the local images onto the global barycenters requires solving the OT problem. \cite{cuturi2013sinkhorn} demonstrates an empirical complexity of $O(d^2)$ with respect to the input dimension $d$. The complexity of our preprocessing is a combination of these two, contingent on the number of agents. Assume we have $N$ agents in our network, each with $M$ images. We must compute $N$ barycenters. Since these are computed in parallel, the time complexity is equivalent to computing one, with the addition of computing the global barycenter. Therefore, we have $O(Md^2/\epsilon^2)$ for local barycenters and $O(Nd^2/\epsilon^2)$ for the global barycenter. The projection time complexity is scaled in accordance with $M$, not $N$, as each agent projects in parallel. Therefore, the overall time complexity for our preprocessing algorithm is $O(Md^2/\epsilon^2 ) + O(Nd^2/\epsilon^2) + O(Md^2)$. If the local number of images vary per agent, the worst-case complexity is the same and requires only setting $M=\text{argmax}_i M_i$ for agents $i=1,\dots, N$.

\section{Conclusion and Future Work}
In this work, we demonstrated the ability of our proposed preprocessing algorithm to improve the convergence speed and generalization of traditional FL. We accomplish this by projecting local data to a space that encodes all local data, in turn minimizing the distributional discrepancy between agents. After aligning the local datasets, we train a model using \textit{FedAvg}. Compared  with certain methods (e.g., FedAvg and MOON), our proposed solution shows higher accuracy scores.  In future work, we will explore larger datasets (e.g., ImageNet) and other modalities. We also aim to develop new OT-based methods for transforming temporal data.

\bibliographystyle{IEEEbib}
\bibliography{refs}

\end{document}